\documentclass[10pt,twocolumn,letterpaper]{article}

\usepackage[pagenumbers]{cvpr} % To force page numbers, e.g. for an arXiv version

\usepackage{graphicx}
\usepackage{amsmath}
\usepackage{amssymb}
\usepackage{booktabs}

\usepackage{pifont}% http://ctan.org/pkg/pifont
\newcommand{\cmark}{\ding{51}\xspace}%
\newcommand{\xmark}{\ding{55}\xspace}%
\usepackage{multirow}

\usepackage[pagebackref,breaklinks,colorlinks]{hyperref}

\usepackage[capitalize]{cleveref}
\crefname{section}{Sec.}{Secs.}
\Crefname{section}{Section}{Sections}
\Crefname{table}{Table}{Tables}
\crefname{table}{Tab.}{Tabs.}

\begin{document}


\title{Integrally Migrating Pre-trained Transformer Encoder-decoders \\for  Visual Object Detection}

\author{Feng Liu\textsuperscript{1}\thanks{Equal Contribution.}\quad 
Xiaosong Zhang\textsuperscript{1}\footnotemark[1]\quad
Zhiliang Peng\textsuperscript{1}\quad
Zonghao Guo\textsuperscript{1}\\
Fang Wan\textsuperscript{1}\quad
Xiangyang Ji\textsuperscript{2}\quad
Qixiang Ye\textsuperscript{1}\\
\textsuperscript{1}University of Chinese Academy of Sciences
\quad
\textsuperscript{2}Tsinghua University\\
{\tt\small liufeng20@mails.ucas.ac.cn}\quad
{\tt\small zhangxiaosong18@mails.ucas.ac.cn}\\
{\tt\small pengzhiliang19@mails.ucas.ac.cn}\quad
{\tt\small guozhonghao19@mails.ucas.ac.cn}\\
{\tt\small wanfang@ucas.ac.cn}\quad
{\tt\small xyji@tsinghua.edu.cn}\quad
{\tt\small qxye@ucas.ac.cn}
}
\maketitle

%%%%%%%%% ABSTRACT
\begin{abstract}
Modern object detectors have taken the advantages of backbone networks pre-trained on large scale datasets. Except for the backbone networks, however, other components such as the detector head and the feature pyramid network (FPN) remain trained from scratch, which hinders fully tapping the potential of representation models.
In this study, we propose to integrally migrate pre-trained transformer encoder-decoders (imTED) to a detector, constructing a feature extraction path which is ``fully pre-trained" so that detectors' generalization capacity is maximized. 
The essential differences between imTED with the baseline detector are twofold: (1) migrating the pre-trained transformer decoder to the detector head while removing the randomly initialized FPN from the feature extraction path; and (2) defining a multi-scale feature modulator (MFM) to enhance scale adaptability. 
Such designs not only reduce randomly initialized parameters significantly but also unify detector training with representation learning intendedly. Experiments on the MS COCO object detection dataset show that imTED consistently outperforms its counterparts by $\sim$2.4 AP. Without bells and whistles, imTED improves the state-of-the-art of few-shot object detection by up to 7.6 AP. 
Code is available at \href{https://github.com/LiewFeng/imTED}{\color{magenta}github.com/LiewFeng/imTED}. 
\end{abstract}

\begin{figure}
  \centering
    \includegraphics[width=1\linewidth]{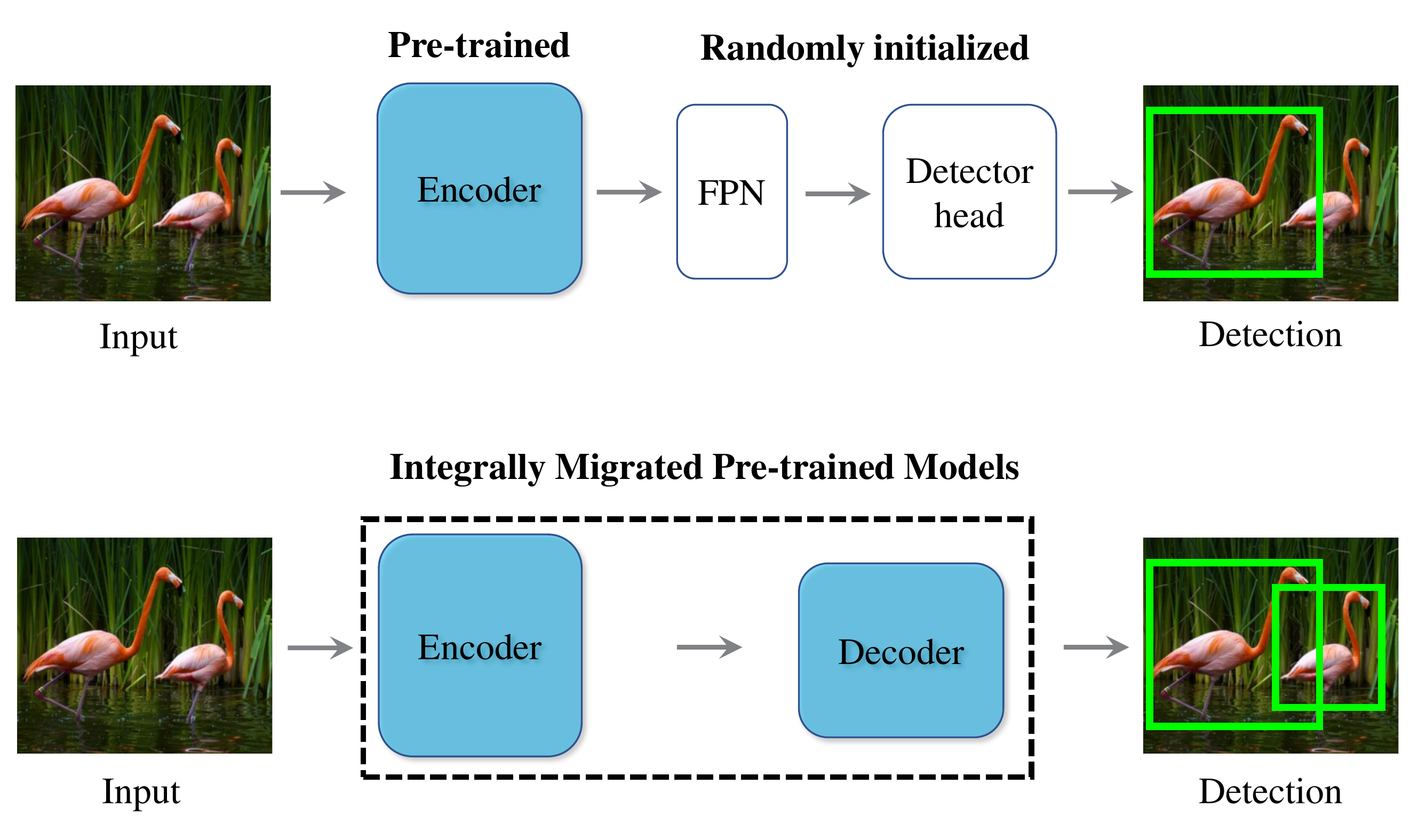}
  \caption{Comparison of the baseline detector $e.g.$, Faster R-CNN~\cite{FasterRCNN2015} equipped with a transformer backbone (\textit{upper}) with the proposed imTED (\textit{lower}). The baseline detector solely transfers a pre-trained backbone network, $e.g.$, the transformer encoder, but training the detector head and FPN from scratch. By contrast, our imTED approach integrally migrates the pre-trained transformer encoder-decoder. It significantly reduces the proportion of randomly initialized parameters and improves detector's generalization capability.}
  \label{fig:introduction}
\end{figure}

%%%%%%%%% BODY TEXT
\section{Introduction}

Over the past two years, vision transformers (ViTs)~\cite{ViT2021} have been promising representation models. The vanilla transformer trained with a sophisticated self-supervised learning method, $e.g.$, masked autoencoder (MAE)~\cite{MAE2021}, demonstrated great potential. Since the introduction of transformers~\cite{Attention2017} to computer vision, the effort of taming them for object detection has never stopped~\cite{UViT2021, benchmarking2021}. This is motivated by the observation that ViTs pre-trained on extraordinarily large-scale datasets incorporate over-completed and versatile features, which guarantee the performance and generalization capability of detectors finetuned on small datasets. ~\cite{benchmarking2021,VitDET2022}. 

Modern object detectors, such as Faster R-CNN and Mask R-CNN~\cite{FasterRCNN2015,MaskRCNN2017}, typically consist of a backbone network, a neck component and a detector head. However, except for the backbone network, other components that occupy a significant proportion of parameters remain trained from scratch, Fig.~\ref{fig:introduction}(upper). Such components, including but not limited to the region proposal network (RPN)~\cite{FasterRCNN2015}, the feature pyramid network (FPN)~\cite{FPN2017} and the detector head~\cite{FastRCNN2015}, fail to take advantages of the representation models pre-trained on large-scale datasets.

In this study, we do not design any new components for object detection; instead, we devote to take full advantages of pre-trained models to improve detector's generalization capability. Specifically, we propose to integrally migrate pre-trained transformer encoder-decoders (imTED) to detectors, Fig.~\ref{fig:introduction}(lower), constructing a feature extraction path which is not only ``fully pre-trained" but also consistent with pre-trained models, as much as possible. 

As shown in Fig.~\ref{fig:introduction}(lower), imTED employs the ViT encoder pre-trained with MAE~\cite{MAE2021} as backbone, and uses the decoder as the detector head. It breaks the routine to remove the randomly initialized FPN from the feature extraction path while leveraging the adaptive respective field provided by the attention mechanism in ViTs~\cite{ViT2021,Conformer2021} to handle objects at multiple scales.
These designs support the integral migration of pre-trained encoder-detector to the object detection pipeline. By adding linear output layers, $i.e.$, a light-weight classification layer and a bounding-box regression layer, atop the migrated encoder-decoder, imTED realizes object classification and localization.
To enhance the capacity for multi-scale object detection, we introduce a multi-scale feature modulator (MFM), which combines both the advantages of FPN with those of fully pre-trained models. 

The competitiveness of imTED is validated upon popular detectors including Faster R-CNN and Mask R-CNN~\cite{FasterRCNN2015, MaskRCNN2017}. Experiments on the MS COCO dataset demonstrate that imTED with ViT-base model outperforms its counterpart by $\sim$2.4 AP at moderate computational cost. Benefiting from the integral migration of pre-trained models, imTED demonstrates strong generalization capability, which is validated by low/few-shot detection tasks. When reducing proportions of the training data, performance gains of imTED monotonously increase. When training a few-shot detector, by freezing the backbone network while finetuning the rest detector components, imTED improves the state-of-the-art by up to 7.6 AP. imTED opens up a promising direction for few-shot object detection using vision transformers.

The contributions of this study include:

\begin{itemize}
    \item We integrally migrate pre-trained transformer encoder-decoders (imTED) to object detectors, constructing a ``fully pre-trained" feature extraction path to improve detectors' generalization capacity.
    
    \item We redesign the feature extraction path to guarantee the ``integral migration" of the pre-trained transformer encoder-decoders. We introduce a multi-scale feature modulator (MFM), to improve the scale adapatiblity of imTED. 
    
    \item imTED not only achieves significant performance gains on object detection and few-shot object detection, but also takes a step towards unifying detector training with representation learning.
\end{itemize}

%-------------------------------------------------------------------------
\section{Related Work}

\textbf{Representation Models.} Object detection has widely explored representation models pre-trained upon large-scale datasets. Over the past decade, CNNs~\cite{krizhevsky2012imagenet, simonyan2014very, szegedy2015going, he2016deep, xie2017aggregated} have been preferred representation models. Recently, vision transformers~\cite{ViT2021, liu2021swin, fan2021multiscale, wang2021pyramid} demonstrated greater potential. Vision transformers including ViT~\cite{ViT2021}, Swin~\cite{liu2021swin}, MViT~\cite{fan2021multiscale}, and PvT~\cite{wang2021pyramid} became promising models for image recognition. The vision transformers~\cite{BEIT2021, zhou2021ibot, beitv2, MAE2021} trained with self-supervised paradigms were validated to have higher generalization capability. Such generalization capability was pushed to a new height by MAE~\cite{MAE2021}, which constructed not only representation models for feature extraction but also decoders for image reconstruction.

Model object detectors, either CNN-based~\cite{liu2016ssd, lin2017focal, FastRCNN2015, FasterRCNN2015, MaskRCNN2017} or transformer-based~\cite{DETR2020, zhu2020deformable}, utilized pre-trained representation models as encoders to extract features, while left the FPN and detector head using randomly initialized parameters. These randomly initialized parameters, when finetuned using few training samples, experience difficult to achieve promising performance. Considering that the backbone, the FPN~\cite{lin2017feature} and the detector head occupy most of the learnable parameters of an object detector, to make them be ``fully pre-trained" is an important problem to be solved.

\textbf{Feature Pyramid Network.} FPN~\cite{lin2017feature} leveraged a top-down structure with lateral connections to construct high-level semantic feature maps at scales, enhancing the flexibility for multi-scale representation. It was designed to adapt hierarchical CNN features but not compatible with plain representation models, $e.g.$, ViT~\cite{ViT2021}. To solve this problem, a small network was designed to obtain multi-scale features~\cite{benchmarking2021}, but this unfortunately caused more parameters being randomly initialized. 

The ViTDet method~\cite{VitDET2022} proposed to remove the top-down feature fusion to simplify FPN, but remains not constructing a ``fully pre-trained" feature extraction path. The major difference between ViTDet~\cite{VitDET2022} and our imTED approach lies in the detector head. imTED simply feeds the last feature map of the MAE encoder to the RoI-Align component, without applying FPN. The aligned features are fed to the pre-trained transformer decoder for object classification and localization. Such designs guarantee that the feature extraction path be consistent with that of the pre-trained model.

\textbf{Detector Head.} DETRs~\cite{DETR2020, zhu2020deformable} are representative detectors, which leverage transformers as the detector head. Given CNN features as input, the transformer encoder-decoder reasons the relations of the objects and the global image context to output the final set of predictions. However, the vision transformers in DETRs were randomly initialized and only used to process features extracted by the backbone network. By contrast, the transformer in our imTED is pre-trained and utilized to not only extract features but also perform feature transformation. As a variant of DETR, ViDT~\cite{song2021vidt} replaced the CNN backbone with a pre-trained transformer but still leaved the following transformer neck randomly initialized. 

Recently, ViTDet~\cite{VitDET2022} and MIMDet~\cite{fang2022unleashing} tried the powerful representations pre-trained by MAE~\cite{MAE2021} for object detection. However, ViTDet solely leverages the pre-trained MAE encoder but deprecates the pre-trained decoder. Whereas, the proposed imTED utilizes both the pre-trained encoder and the pre-trained decoder. 
Although MIMDet~\cite{fang2022unleashing} utilizes both the encoder and decoder for feature extraction, the core idea is leveraging the reconstruction ability of  decoder to mask input image patches, which reduces the computation cost. It keeps the randomly initialized FPN and detector head, as well as introducing more randomly initialized layers for multi-scale feature extraction. By contrast, the imTED approach in this study utilizes the pre-trained encoder to extract features and the pre-trained decoder as the detector head, constructing a ``fully pre-trained" feature extraction path, for the first time to our best knowledge.

%-------------------------------------------------------------------------
\section{Approach}

The goal of this study is to integrally migrate the pre-trained transformer encoder-decoder as the pillars of an object detector. To this end, we choose encoder-decoders pre-trained by MAE~\cite{MAE2021} and migrate them to conventional two-stage detectors, $e.g.$, Faster R-CNN and Mask R-CNN~\cite{FasterRCNN2015,MaskRCNN2017}. In what follows, we first describe the motivation of imTED. We then address how to integrally migrate the pre-trained encoder-decoders. Finally. we describe the implementation details of an imTED detector. We also show that modulating multi-scale features to the fully pre-trained feature extraction path further boosts the detection performance. 

\begin{figure*}
  \centering
  \includegraphics[width=1.0\linewidth]{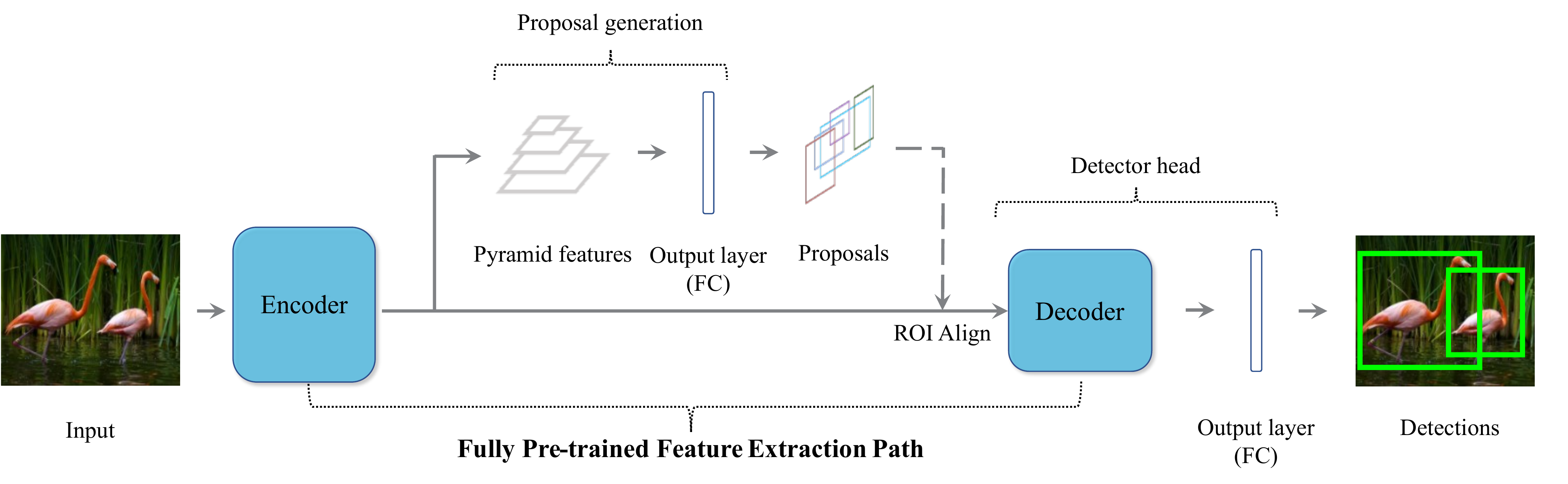}
  \caption{Architecture of the imTED detector. By integrally migrating the transformer encoder-decoder, imTED constructs a feature extraction path, which is ``fully pre-trained". The reconstructed feature pyramid is only applied for object proposal generation but does not involve the feature extraction procedure. With these designs, the proportion of randomly initialized network parameters of the detector is significantly reduced.
  }
  \label{fig:arc}
\end{figure*}

\subsection{Motivation}
MAE pre-trains encoder-decoder representation models based on the pretext task of masked image modeling~\cite{MAE2021}. 
By randomly masking image patches and reconstructing the masked patches, it trains an encoder for feature extraction and a decoder for image context modeling.
It was validated that the MAE decoder has the ability to reconstruct masked pixels under a high mask ratio of 75\%~\cite{MAE2021}, demonstrating strong capacity to model image context information.
This piques our curiosity: \textit{could the spatial context modeling capacity of the MAE decoder benefits object localization?}

To answer this question, we conduct an experiment about single object detection on the ImageNet Localization Dataset~\cite{ImageNet2009}\footnote{Please refer to the supplementary material for details of dataset preparation.}
In the experiment, detectors are trained to predict a single object in each image to avoid the interference of complicated feature-object matching, design of FPN, and/or RoI alignment.
Three variants of the object feature extractor are compared: (i) pre-trained encoder only; (2) pre-trained encoder with randomly initialized decoder; (3) pre-trained encoder with pre-trained decoder (imTED).
Following the feature extractor, an object localization head and a classification head is used to realize object detection. 

As shown in Table \ref{tab:localization}, the introduction of the randomly initialized decoder boosts the detection performance by 0.5 mAP and the localization performance by 0.2 CoLoc.
Go a step further, the pre-trained decoder improves the detection performance by 1.4 mAP and the localization performance by 0.9 CoLoc. 
The significant performance gains validate that \textit{the context modeling capacity of the pre-trained decoder does benefit object localization,} which motivates our integral migration approach.

\begin{table}[t]
\caption{Object detection and localization performance under three decoder variants on the ImageNet Localization Dataset. mAP and CoLoc are calculated under 0.5 IoU. CoLoc measures the correctly localized object ratio.}
\label{tab:localization}
\centering
\resizebox{\linewidth}{!}{ %resize
\begin{tabular}{lccc}
\toprule
Model Variants          & mAP & CoLoc & Acc.   \\
\midrule
pre-trained encoder    & 43.4 & 77.4 & 77.1   \\
~~$+$ random decoder  & 43.9 (+0.5) & 77.6 (+0.2) & 77.7 (+0.6)   \\
~~$+$ \bf pre-trained decoder & 44.8 (+\bf1.4) & 78.3 (+\bf0.9) & 78.0 (+\bf0.9)   \\
\bottomrule
\end{tabular}
} %resize
\end{table}

\subsection{Constructing A Fully Pre-trained Feature Extraction Path}
\label{sec.path}

\textbf{Baseline Detectors.} The Faster R-CNN~\cite{FasterRCNN2015} and Mask R-CNN~\cite{MaskRCNN2017} are employed as baseline detectors. The detector mainly consists of four components: a backbone network, a feature pyramid network (FPN), a region proposal network (RPN) and a detector head. By adding a mask head atop Faster R-CNN, Mask R-CNN can simultaneously conduct object detection and instance segmentation. The components of a conventional detector are partially pre-trained. The backbone network is pre-trained on large-scale datasets, while the FPN, RPN and detector head, which occupy a large proportion ($\sim 40\%$) of learnable parameters, are trained from scratch. The reason to use randomly initialized components lies in that the backbone networks specified for image classification~\cite{ImageNet2009} can not be directly applied for multi-scale feature extraction and object localization.

\textbf{Integral Migration of Encoder-Decoder.} As shown in Fig.~\ref{fig:arc}, we redesign the feature extraction path  by integrally migrating the transformer encoder and decoder pre-trained with MAE. The created imTED detector not only leverages the encoder for feature extraction but also the decoder for feature transformation. It then leverages a fully connected layer, a light-weight layer, for object classification and localization. Notice that the proposal generation pipeline remains unchanged, $i.e.$, the FPN and RPN remain using randomly initialized parameters. Whereas, the proposal generation pipeline is only responsible for producing region of interests (RoIs) but does not involve object feature extraction or transformation. Thereby, the randomly initialized parameters would not deteriorate detector's generalization capacity.

With these redesigns, the imTED detector has significantly fewer parameters trained from scratch, mostly lie in the proposal generation path, Fig.~\ref{fig:arc}. When using the ViT-S~\cite{ViT2021} model, for example, the Faster R-CNN detector has $\sim$17.7M parameters trained from scratch, while imTED changes this figure to $\sim$3.3M, which infers a reduction of 81.3\%. As is known, larger proportions of pre-trained parameters imply higher generalization capability. imTED thereby enjoys significantly higher performance than the baseline detector.

\begin{figure*}
  \centering
  \begin{subfigure}[b]{0.31\linewidth}
    \centering
    \includegraphics[width=0.95\linewidth]{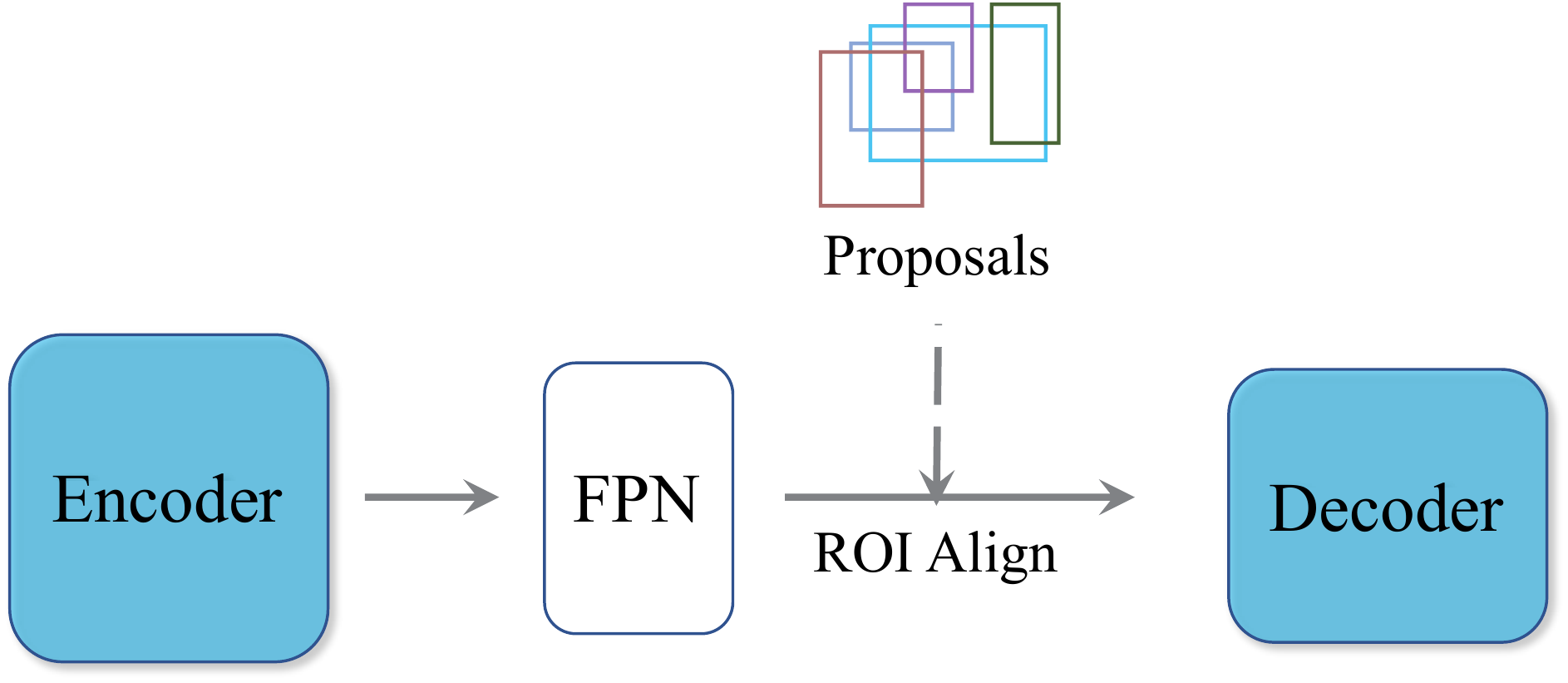}
    \caption{Faster R-CNN}
  \end{subfigure}
  \begin{subfigure}[b]{0.31\linewidth}
    \centering
    \includegraphics[width=0.95\linewidth]{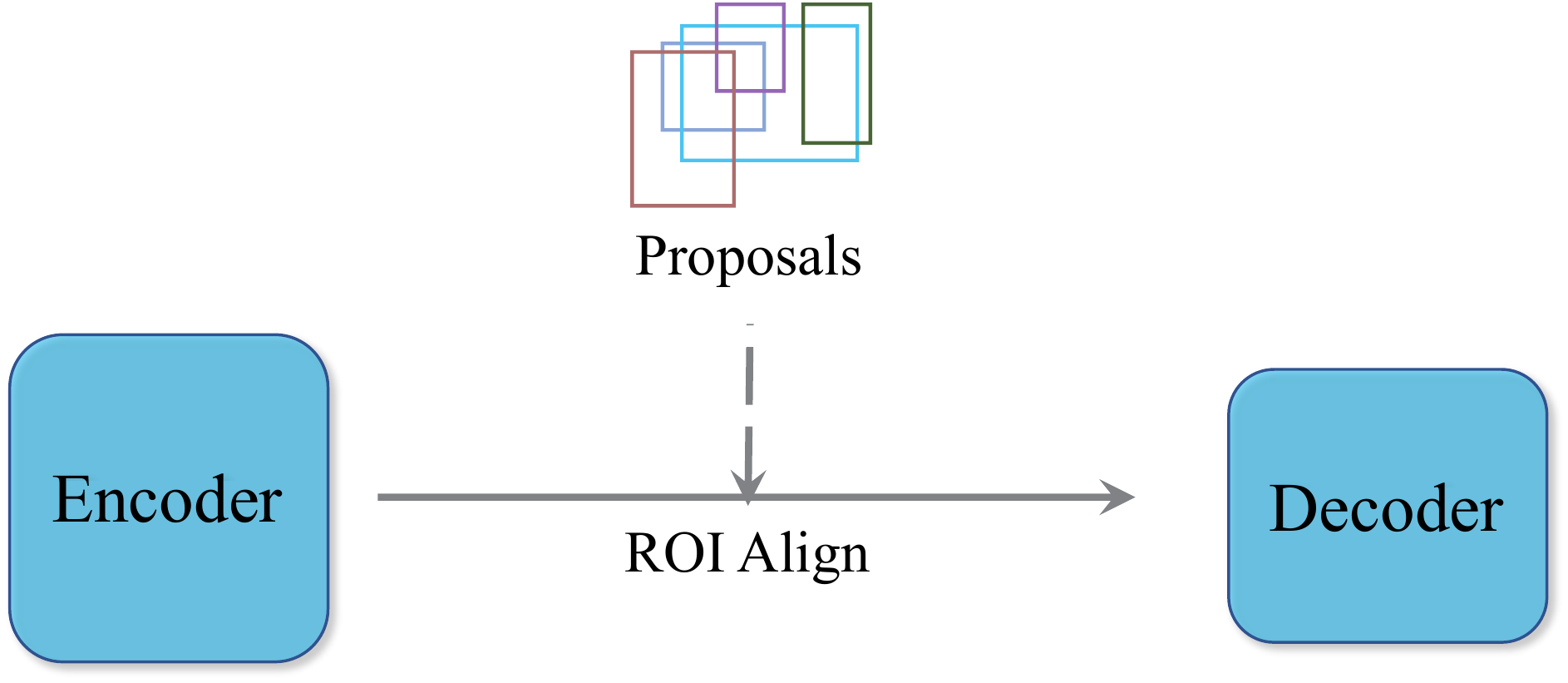}
    \caption{imTED-SS}
  \end{subfigure} 
  \begin{subfigure}[b]{0.37\linewidth}
    \centering
    \includegraphics[width=0.95\linewidth]{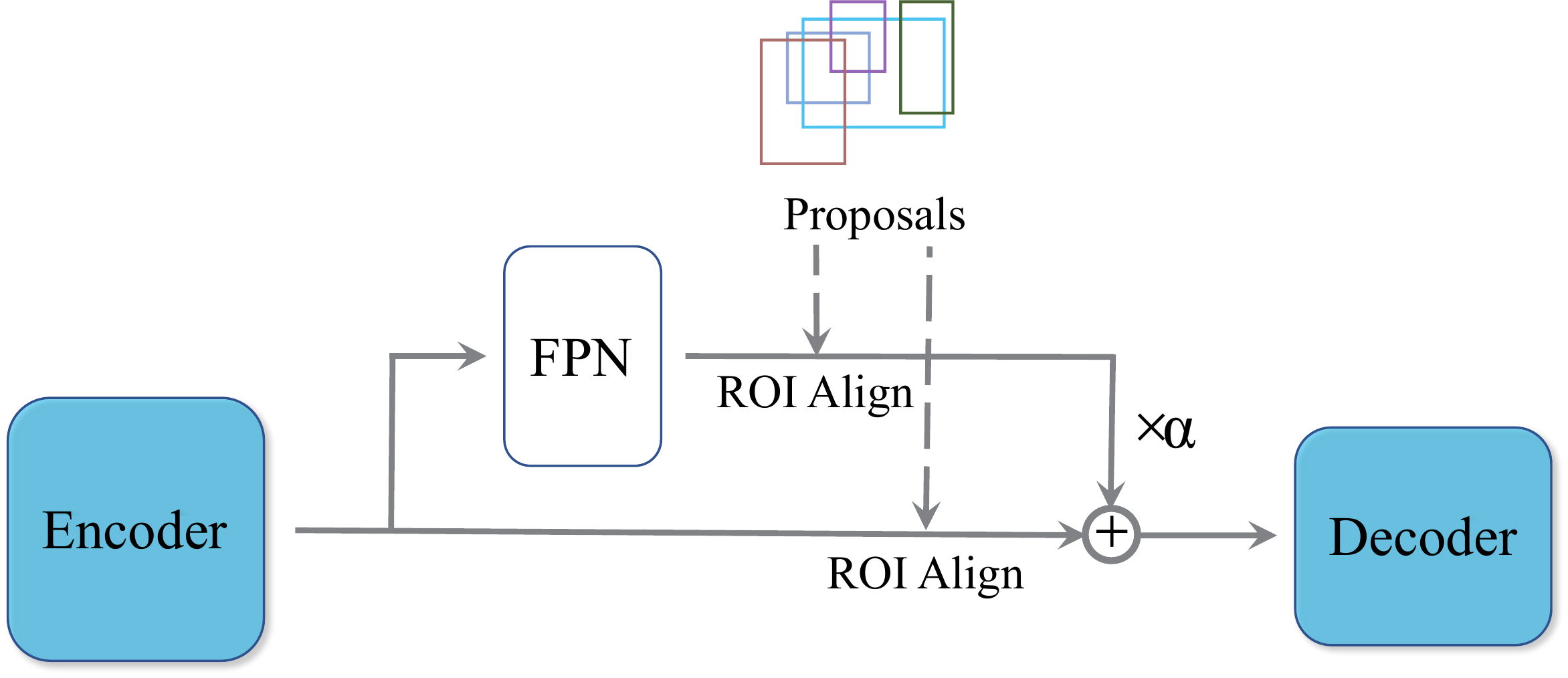}
    \caption{imTED}
  \end{subfigure}
  \caption{The involvement of (a) a conventional Faster R-CNN detector to (b) a single-scale detector (imTED-SS) and (c) the imTED detector with multi-scale feature modulating. 
  }
  \label{fig:architecture}
\end{figure*}

\textbf{Removing Feature Pyramid Network.} 
In Faster R-CNN, FPN can be deployed atop the encoder to augment the features to multiple resolutions~Fig.~\ref{fig:architecture}(a). With FPN, large objects are represented by the low-resolution features and small ones by high-resolution features. However, FPN is constructed by using randomly initialized parameters, which violates the ``fully pre-trained" idea. Fortunately, benefiting from the global attention mechanism, the transformer encoder is able to construct an adaptive receptive field~\cite{Conformer2021}, which reduces the requirement of scale alignment between objects and features.
As a result, we are able to remove the FPN from the feature extraction path, Fig.~\ref{fig:architecture}(b). It is no doubt that removing FPN has a negative impact on the multi-resolution representation capability of features. Nevertheless, significant performance gains are observed in experiments, which supports the idea that constructing a ``fully pre-trained" feature extraction path is more important than the multi-scale prior.

\subsection{Detector Implementation}
As described in Sec.~\ref{sec.path}, by migrating the transformer encoder as the backbone, plugging the decoder to the detector head, and removing the FPN, we construct a ``fully pre-trained" feature extraction path. The architecture of RPN is not updated as it plays the role of generating region proposals but does not disturb the feature extraction stream. An imTED detector is then implemented by simply adding a few linear layers and a proposal generation module to the fully pre-trained encoder-decoder, Fig.~\ref{fig:arc}.

\textbf{Backbone Network.}  There is no modification to the transformer encoder except for resizing the encoder's positional embeddings so that they are consistent with input image sizes. The transformer encoder, pre-trained on a large-scale dataset, outputs a single-scale feature map which is down-sampled by a factor of 16 relative to the input image. The single-scale feature map is fed to the RoI-Align module for proposal feature extraction.

\textbf{Region Proposal Generation.}
In the two-stage detection architecture~\cite{FasterRCNN2015}, dense and multi-scale region proposals are used for object localization. 
% To produce multi-scale proposals, following~\cite{VitDET2022}, we make multi-scale feature maps by placing four down-sampling and de-convolution layers atop the encoder. 
To produce multi-scale feature maps, we up-sample or down-sample intermediate ViT feature maps by placing four resolution-modifying modules at equally spaced intervals of $d$/4 transformer blocks following~\cite{benchmarking2021}, where $d$ denotes the total number of blocks. 
The multi-scale feature maps are fed to the FPN, the output of which is further fed to the RPN for proposal generation. The training of the RPN parameters, $i.e.$, the weights of the fully connected output layer, is consistent with that of Faster R-CNN~\cite{FasterRCNN2015}. 

\textbf{Detector Head.} A pre-trained MAE decoder is migrated to the detector head to replace the randomly initialized network parameters, Fig.~\ref{fig:arc}. The detector head consists of the pre-trained decoder and two linear layers. Given the feature map extracted by the encoder, an RoI-Align module is used to extract features for each region proposal. The extracted features are then embedded with location information by summarizing with position embeddings~\cite{MAE2021}. The features with position embedding are then fed to the decoder and transformed with alternative attention and MLP layers. The transformed features are finally fed to the linear classification and regression layers to predict object categories and location offsets.

\subsection{Multi-scale Feature Modulator}
\label{sec.fp}

Although the single-scale feature extracted by the transformer encoder is adaptive to object scales to some extent, we are wondering could the multi-scale feature representation be recalled back, in a new fashion, to further enhance the scale adaptability? To defend the idea of ``fully pre-training", we can not directly call the FPN back to the feature extraction path; instead we redefine FPN as a multi-scale feature modulator (MFM), which acts after the RoI-Align module~Fig.~\ref{fig:architecture}(c). Feature modulation for region proposals is defined as an adaptive linear weighting procedure, as   
\begin{equation}
    \label{eq:fusion}
        \mathbf{F}=\mathbf{F}_{ss} + \boldsymbol{\alpha} * \mathbf{F}_{ms},
\end{equation}
where $\mathbf{F} \in \mathbb{R}^{C\times H\times W}$ denotes the weighted features. $\mathbf{F}_{ss} \in \mathbb{R}^{C\times H\times W}$ denotes single-scale feature extracted by the pre-trained encoder and $\mathbf{F}_{ms} \in \mathbb{R}^{C\times H\times W}$ denotes multi-scale features extracted by the randomly initialized FPN, which is constructed by using the single-scale feature as input~\cite{FPN2017}.
$\boldsymbol{\alpha} \in \mathbb{R}^{C}$ is a learnable weight vector. $H$ and $W$ respectively denote the height and weight of the output feature maps of the RoI Align module~\cite{FasterRCNN2015}. Both $H$ and $W$ are set to 7, following the setting of Faster R-CNN~\cite{FasterRCNN2015}. $C$ is the channel dimension. 

At the start-point of detector training, the elements of $\boldsymbol{\alpha}$ in Eq.~\ref{eq:fusion} are initialized to zeros. When detector training proceeds, $\boldsymbol{\alpha}$ gradually updates so that the single-scale feature extracted by encoder is adaptively combined with the multi-scale features. In a learnable way, the multi-scale representation capacity is modulated to the single-scale representation. The evolution of Faster R-CNN to imTED-SS and imTED is illustrated in Fig.~\ref{fig:architecture}.

%-------------------------------------------------------------------------
\section{Experiment}

\label{sec: exp}
\def \mark {$^{\star}$}
\def \bbox {$^{\text{box}}$}
\def \mask {$^{\text{mask}}$}
\def \50 {$_{50}$}
\def \75 {$_{75}$}
\def \S  {$_{\text{S}}$}
\def \M  {$_{\text{M}}$}
\def \L  {$_{\text{L}}$}
\def \maskS  {$^{\text{mask}}_{\text{S}}$}
\def \maskM  {$^{\text{mask}}_{\text{M}}$}
\def \maskL  {$^{\text{mask}}_{\text{L}}$}

\subsection{Setting}
The ViT models are categorized to ViT-S, ViT-B and ViT-L~\cite{ViT2021} according to the parameter scales. These models are pre-trained on ImageNet-1K using the self-supervised MAE method~\cite{MAE2021} for 1600 epochs. By adding a proposal generation module, RoI-Align module, multi-scale feature modulator and light-weight linear output layers atop the pre-trained encoder-decoder, the imTED detector is constructed. 
The detectors are evaluated on the MS COCO dataset~\cite{COCO2014}, which consists of $\sim$118k training images and 5k validation images. Data augmentation strategies are defined by resizing image with shorter size between 480 and 800 while the longer side is no larger than 1333~\cite{DETR2020} . 
The detector is trained using the AdamW optimizer~\cite{AdamW2018} with a learning rate 1e-4, a weight decay of 0.05. The training lasts for 3$\times$schedule (36 epochs with the learning rate decayed by 10 at epochs 27 and 33). The batch size is 16, distributed across 8 GPUs (2 images per GPU). For the ViT-S/B/L models, a layer-wise lr decay~\cite{BEIT2021} of 0.75 and a drop path rate of 0.1/0.2/0.3 are also applied.  

\begin{table*}
  \caption{Object detection performance on the MS COCO dataset. Comparison of the proposed imTED detector with the state-of-the-art detectors using vision transformers as backbones. None of compared detection methods (ViTDet, MIMDet, imTED-SS and imTED) uses relative position embedding.}
  \label{tab:main}
  \centering
  \begin{tabular}{lllcccccc}
    \toprule
    \multirow{2}*{Approach} & \multirow{2}*{Backbone}    & \multirow{2}*{Pre-train}     & \multirow{2}*{Epochs}
     & \multicolumn{3}{c}{\small Faster R-CNN} & \multicolumn{2}{c}{\small Mask R-CNN}                     \\
    \cmidrule(r){5-9}
      \multicolumn{4}{c}{}  &AP    &  AP\50     & AP\75      & AP\bbox    & AP\mask     \\
    \midrule
    Baseline~\cite{xie2017aggregated}        
                   & ResNeXt101      &  1k, sup     & 36          & 43.1       & 63.6         & 47.2          & 44.5       & 39.7        \\
    Baseline~\cite{liu2021swin}        
                   & Swin-B      &  1k, sup      & 36          & -       & -         & -          & 48.5       & 43.4        \\
    Baseline~\cite{li2021improved}      
                   & MViTv2-B    &  1k, sup      & 36          & -       & -         & -          & 51.0       & 45.7        \\
   Baseline~\cite{FasterRCNN2015}   
                   & ViT-B       & 1k, MAE      & 36          & 50.5    & 71.4      & 55.5      & 51.3       &  45.3\\
    \hline
    ViT-Adapter~\cite{chen2022vision}      
                   & ViT-S       & 1k, sup      & 36         & -       & -         & -          & 48.2       & \bf42.8        \\
    imTED-SS(ours)   
                   & ViT-S       & 1k, MAE      & 36          & 47.3    & 68.6      & 51.0       & 48.0        & 42.4        \\
    % imTED-FPN (ours)   
    %               & ViT-S       & 1k, MAE  & \cmark     & 36          & 48.3     & 68.5       & 52.4       &        &         \\
    imTED(ours)   
                   & ViT-S       & 1k, MAE      & 36          & \bf48.2     & \bf68.4       & \bf52.6       & \bf48.7       & 42.7        \\
    \hline
    ViT-Adapter~\cite{chen2022vision}      
                   & ViT-B       & 1k, sup      & 36         & -       & -         & -          & 49.6       & 43.6        \\
    Li et al.~\cite{benchmarking2021}      
                   & ViT-B       & 1k, MAE       & 100         & -       & -         & -          & 50.3       & 44.9        \\
    ViTDet~\cite{VitDET2022}      
                   & ViT-B       & 1k, MAE       & 100         & -       & -         & -          & 51.6       & 45.9        \\
    MIMDet~\cite{fang2022unleashing}      
                   & ViT-B       & 1k, MAE      & 36          & -       & -         & -          & 51.7       & 46.1        \\
    imTED-SS(ours)   
                   & ViT-B       & 1k, MAE       & 36          & 52.2    & 72.8      & 57.1       & 52.3       & 46.0        \\
    
    imTED(ours)   
                   & ViT-B       & 1k, MAE       & 36          & \bf52.9    & \bf73.2      & \bf57.9       & \bf53.3       & \bf46.4        \\
    \hline
    ViT-Adapter~\cite{chen2022vision}      
                   & ViT-L       & 22k, sup       & 36         & -       & -         & -          & 52.1       & 46.0        \\
    Li et al.~\cite{benchmarking2021}      
                   & ViT-L       & 1k, MAE       & 100         & -       & -         & -          & 53.3       & 47.2        \\
    ViTDet~\cite{VitDET2022}      
                   & ViT-L       & 1k, MAE       & 100         & -       & -         & -          & 55.1       & \bf48.9        \\
    
    MIMDet~\cite{fang2022unleashing}      
                   & ViT-L       & 1k, MAE       & 36          & -       & -         & -          & 54.3       & 48.2        \\
    imTED(ours)   
                   & ViT-L       & 1k, MAE       & 36          & \bf55.4    & \bf75.4      & \bf60.6       & \bf55.5       & 48.1        \\
    \bottomrule
  \end{tabular}
\end{table*}

\begin{table*}
  \caption{Ablation studies using ViT-S as the backbone (encoder) in 1x schedule. \mark~indicates that the module is initialized using MAE pre-trained weights.}
  \label{tab:head}
  \centering
  \begin{tabular}{lccccccccccc}
    \toprule
    Detector Head    & FPN  &MFM      & Params    & FLOPs          & AP       & AP\50     & AP\75     & AP\S     & AP\M     & AP\L        \\
    \midrule
    % FC Layers         & \xmark   &  \cmark   & 41.1M     & 343G            & 41.2     & 62.5      & 44.8      & 24.1     & 44.2     & 55.1        \\
    Conv Layers   &  \cmark  &\xmark      & 42.6M     & 403G           & 42.4      & 62.9     & 46.0      & 25.3      & 45.5     & 56.4           \\
    Decoder        &  \cmark &\xmark       & 30.1M     & 415G            & 42.2     & 62.4      & 45.8      & 25.8     & 45.0     & 57.0        \\
    Decoder\mark     &  \cmark &\xmark      & 30.1M     & 415G           & 42.5     & 63.0      & 46.1      & 25.6     & 45.4     & 57.7        \\
    %
    %Decoder\mark    & \cmark &\xmark   & 30.1M     & 415G            & 43.1     & 64.1      & 46.9      & 25.0     & 46.5     & 58.6        \\
    %
    Decoder\mark    & \xmark &\xmark   & 30.1M     & 415G            & 43.2     & 63.9      & 46.9      & 25.0     & 46.6     & 58.6        \\
    % Decoder\mark     & \checkmark & \checkmark     & 33.6M     & 564G            & 44.2     & 64.4      & 48.0      & 26.1     & 47.5     & 59.2        \\
    Decoder\mark & \xmark &\cmark    & 30.3M     & 430G            & 44.0      & 64.6      & 47.6      & 26.2     & 47.3     & 59.3        \\
    %Decoder\mark     & \cmark & \cmark     & 30.3M     & 366G            & 43.0      & 63.5      & 46.4      & 25.6     & 46.2     & 57.7        \\
    \bottomrule
  \end{tabular}
\end{table*}

\subsection{Detection Performance}

In Table~\ref{tab:main}, imTED detectors are evaluated and compared with the baseline and state-of-the-art detectors. By replacing the ResNeXt101 backbone with a pre-trained ViT model, the baseline detector improves the average precision (AP) from 43.1 to 50.5, setting a solid baseline. Upon the solid baseline, imTED-SS with ViT-B model improves the AP by 1.7 (from 50.5 to 52.2), which is a large margin for the challenging task. Note that this improvement is achieved without using FPN in the feature extraction path, which substantially validates the ``integral migration" idea. When using multi-scale feature modulation (MFM), the total performance gain increases to 2.4 (52.9 \vs 50.5). 
imTED respectively improves the AP$_{50}$ by 1.8 (from 71.4 to 73.2), and the AP$_{75}$ by 2.4 (from 55.5 to 57.9). When using the Mask R-CNN framework, it respectively improves the AP$^{box}$ by 2.0 and the AP$^{mask}$ by 1.1, which are all significant margins. imTED also significantly outperforms the state-of-the-art detectors, $i.e.$, MIMDet and ViTDet, which use pre-trained transformers as backbones. ViTDet solely leverages the pre-trained encoder but deprecates the decoder. 
MIMDet leverages both the encoder and decoder for feature extraction but remains using a randomly initialized detector head, which deteriorates its generalization capability. imTED overcomes these disadvantages and achieves higher performance. Without using MFM, the AP$^{box}$ and AP$^{mask}$ of imgTED-SS respectively outperform the ViTDet detector (which uses FPN) by 0.7 (52.3 \vs 51.6) and 0.1 (46.0 \vs 45.9). When using 
MFM, the improvements of AP$^{box}$ and AP$^{mask}$ rise up to 1.7 and 0.5. 
When using the large backbone (ViT-L), the AP$^{box}$ and AP$^{mask}$ of imTED respectively outperform MIMDet by 1.1 and 0.9. Note that even only trained for 36 epochs, the imTED is comparable to, if not outperforms, ViTDet which is trained for 100 epochs.

\subsection{Ablation Study}

In ablations, we fine-tune the detector for 1× schedule (12 epochs with the learning rate decayed by 10× at epochs 9 and 11) on the \textit{train}2017 split and evaluate on the \textit{val}2017 split. By default, the ViT-S~\cite{DeiT2021} is set as the backbone (encoder), and a 4-layer decoder with 256 dimensions is employed as the detector head. Unless otherwise specified, the ablation experiments are performed on Faster R-CNN.

\textbf{Integral Migration.} The baseline detector (Faster R-CNN) only uses a pre-trained encoder as backbone following~\cite{benchmarking2021}. Its predictions are obtained from FPN, convolutional (Conv) and fully connected layers in the detector head. By replacing the Conv layers in the detector head with the pre-trained MAE decoder and removing the FPN, we construct an integrally pre-trained feature extraction path.
In Table~\ref{tab:head}, when replacing Conv layers in the detector head with a decoder without pre-training, there is a little performance drop -0.2 (42.2 \vs 42.4) observed. When using the encoder as the backbone and the decoder pre-trained by the MAE as the detector head, the AP performance improves 0.3 (42.5 \vs 42.2). 

\textbf{Removing FPN.} By skipping FPN and constructing a fully pre-trained feature extraction path, imTED further improves AP by 0.7 (43.2 \vs 42.5). The total performance gain (43.2 \vs 42.4) over the baseline detector, considering the extensively investigated problem and the challenging aspects of the dataset, validates the effectiveness of the proposed imTED approach.

\begin{figure}
  \centering
  \includegraphics[width=0.47\linewidth, height=0.5\linewidth]{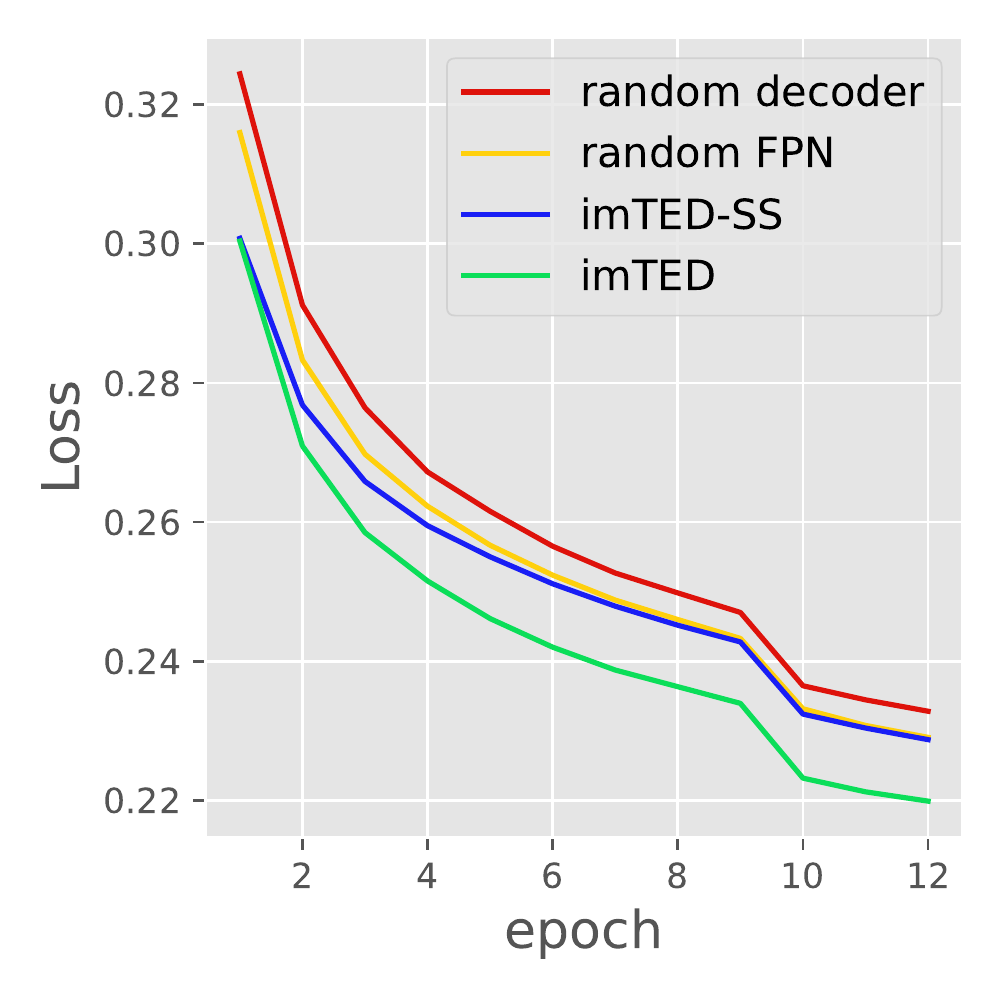} 
  \quad
  \includegraphics[width=0.47\linewidth, height=0.5\linewidth]{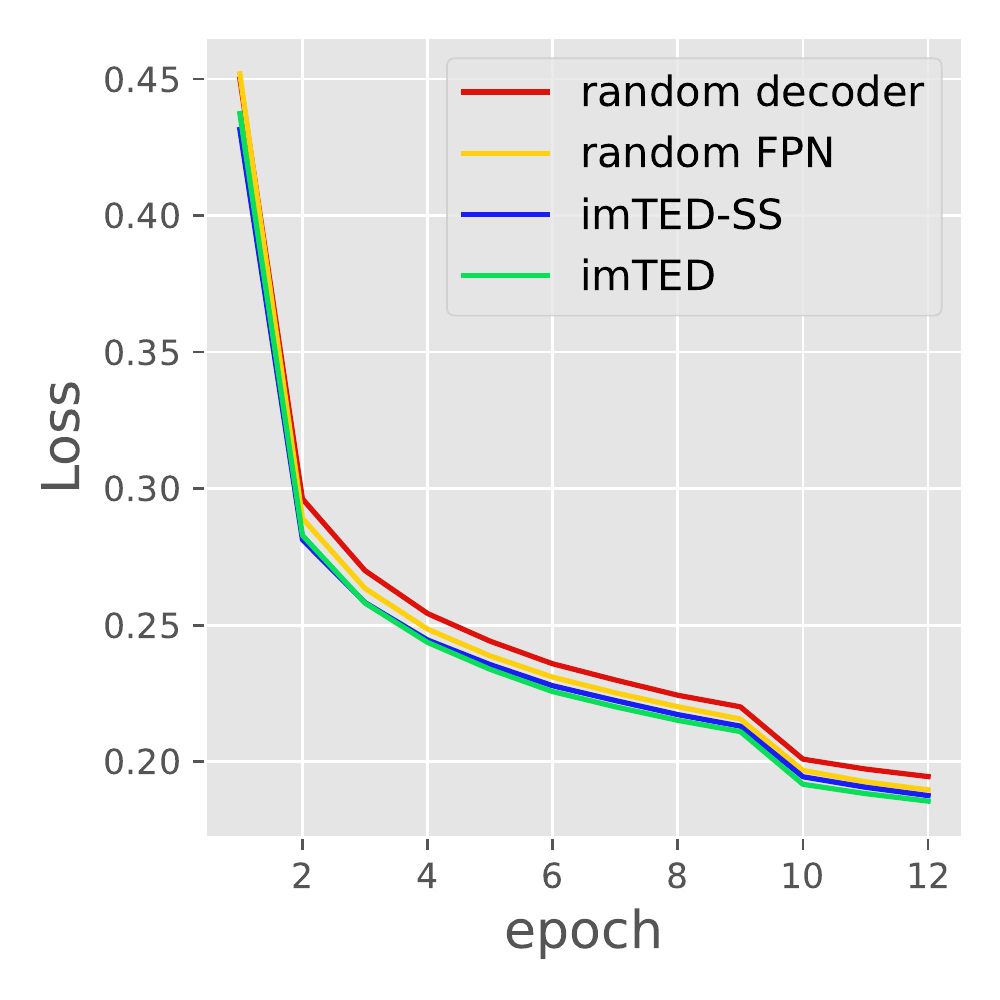}
  \caption{Comparison of object localization loss (left) and object classification loss (right).}
  \label{fig:loss}
\end{figure}

\begin{figure}
  \centering
  \includegraphics[width=0.47\linewidth, height=0.5\linewidth]{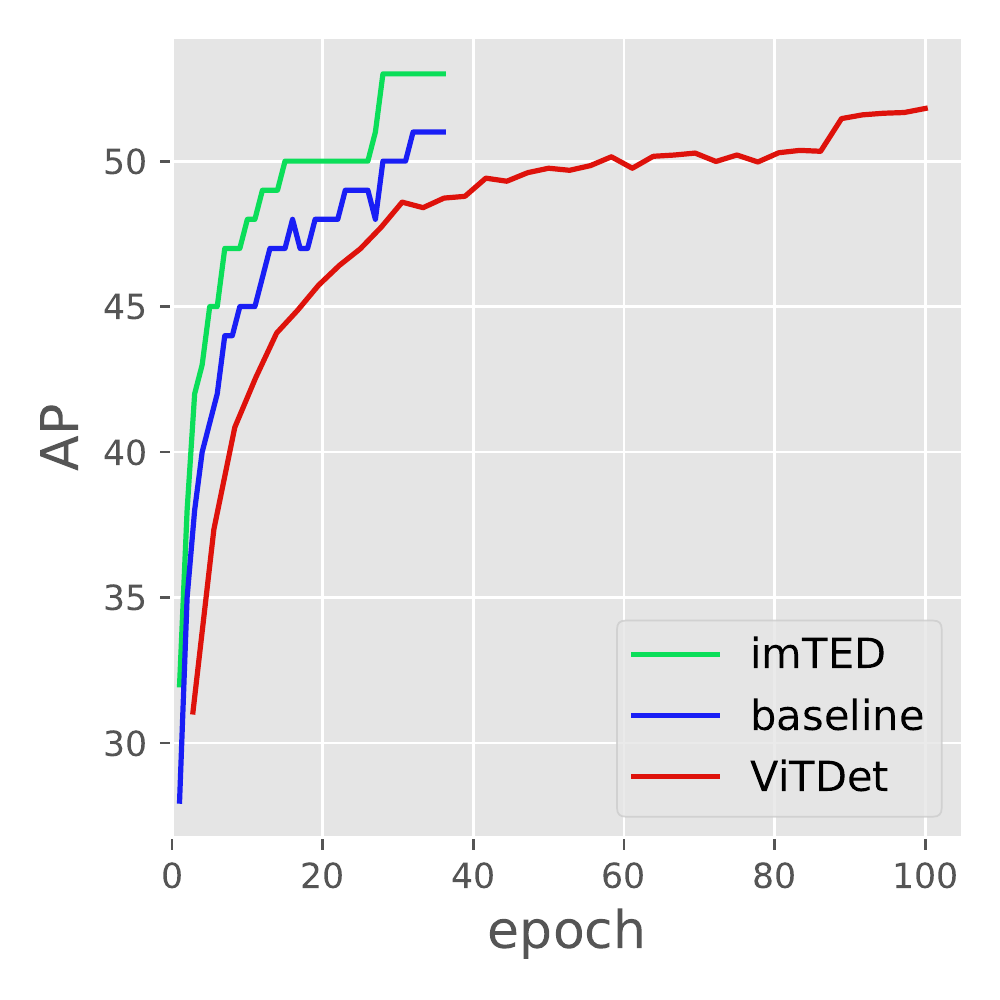} 
  \quad
  \includegraphics[width=0.47\linewidth, height=0.5\linewidth]{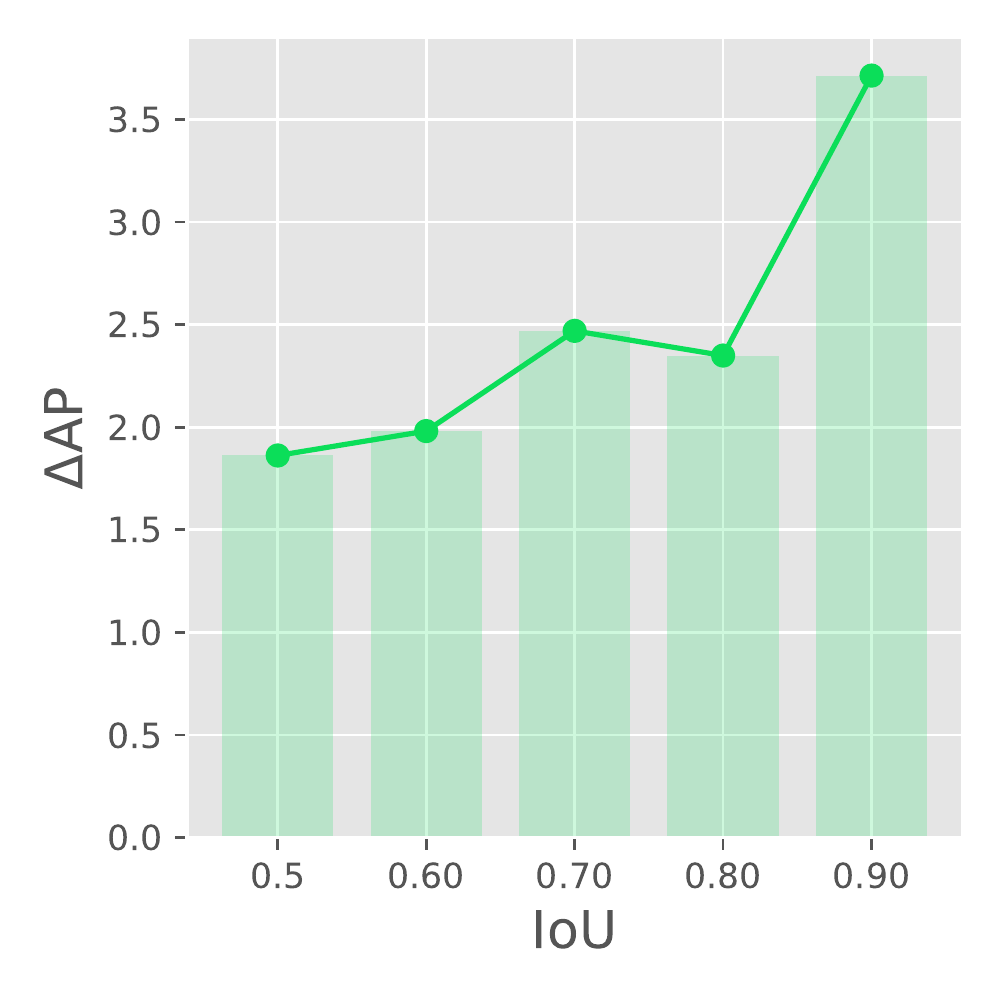}
  \caption{Performance gains during a 3$\times$ training with ViT-B. Left: AP improvements. Right: AP improvements under different IoU thresholds.}
  \label{fig:training_gains}
\end{figure}

\textbf{MFM.} In Table~\ref{tab:head}, when using multi-scale features to modulate the feature extracted by fully pre-trained models, imTED improves AP performance by 0.8 (44.0 \vs 43.2). This shows the compatibility of fully pre-trained models with the randomly initialized module. In total, imTED improves the AP performance by 1.6 (44.0 \vs 42.4).

\textbf{Training Loss Analysis.} As shown in Fig.~\ref{fig:loss}(left), imTED's localization loss decreases faster than the baseline detector using either a randomly initialized decoder or a randomly initialized FPN. imTED benefits from both the integral migration and multi-scale feature representation, demonstrating larger advantages on the localization ability. On the other hand, the compared detectors have similar classification loss curves, Fig.~\ref{fig:loss}(right). This shows that imTED benefits a lot from the strong localization capacity of the pre-trained decoder. 

\begin{table}[t]
  \caption{
    Detection performance using ViT-S in 1x schedule under different detector head depth.}
  \label{tab:bbox head depth}
  \centering
   \resizebox{\linewidth}{!}{
  \begin{tabular}{cccccccccc}
    \toprule
    Depth      & FLOPs         & AP       & AP\50     & AP\75     & AP\S     & AP\M    & AP\L        \\
    \midrule
    1      & 371G               & 42.3     & 62.8      & 45.8      & 25.1     & 45.4     & 56.5        \\
    2      & 390G             & 43.1     & 63.4      & 46.8      & 26.1     & 46.3     & 57.4        \\
    3      & 410G        & 43.9     & 64.2      & 47.3      & 26.0     & 46.9      & 58.7         \\
    4      & 430G       & 44.0      & 64.6      & 47.6      & 26.2     & 47.3     & 59.3        \\
    \bottomrule
  \end{tabular}
  }
\end{table}

\textbf{Depth of Decoder.} In Table~\ref{tab:bbox head depth}, we evaluate the effect of depth of decoder (transformer blocks). Performance gradually saturates when the depth of decoder increases and the 4-layer decoder achieves the best performance. While larger objects benefit more from deeper decoders than smaller ones, the computational cost increases with the number of decoder layers.

\textbf{Performance Gains During Training.}
The imTED detector significantly improves AP during training, Fig.~\ref{fig:training_gains}, which demonstrates that imTED not only speeds up the convergence of training but also raises the performance upper-bound.
Particularly, imTED achieves larger performance gains under larger IoU thresholds, which implies improved localization capacity. 

\textbf{Computational Cost.} In Table~\ref{tab:head}, replacing the Conv layers with the pre-trained decoder brings moderate increase of computational cost, $i.e.$, the FLOPs increases from 403G to 415G. When introducing MFM as the modulator, the FLOPs further increases from 415G to 430G. In total, the FLOPs increase by 6.7\%.

\subsection{Generalization Capacity}

\begin{figure}[t]
    \includegraphics[width=0.47\linewidth, height=0.5\linewidth]{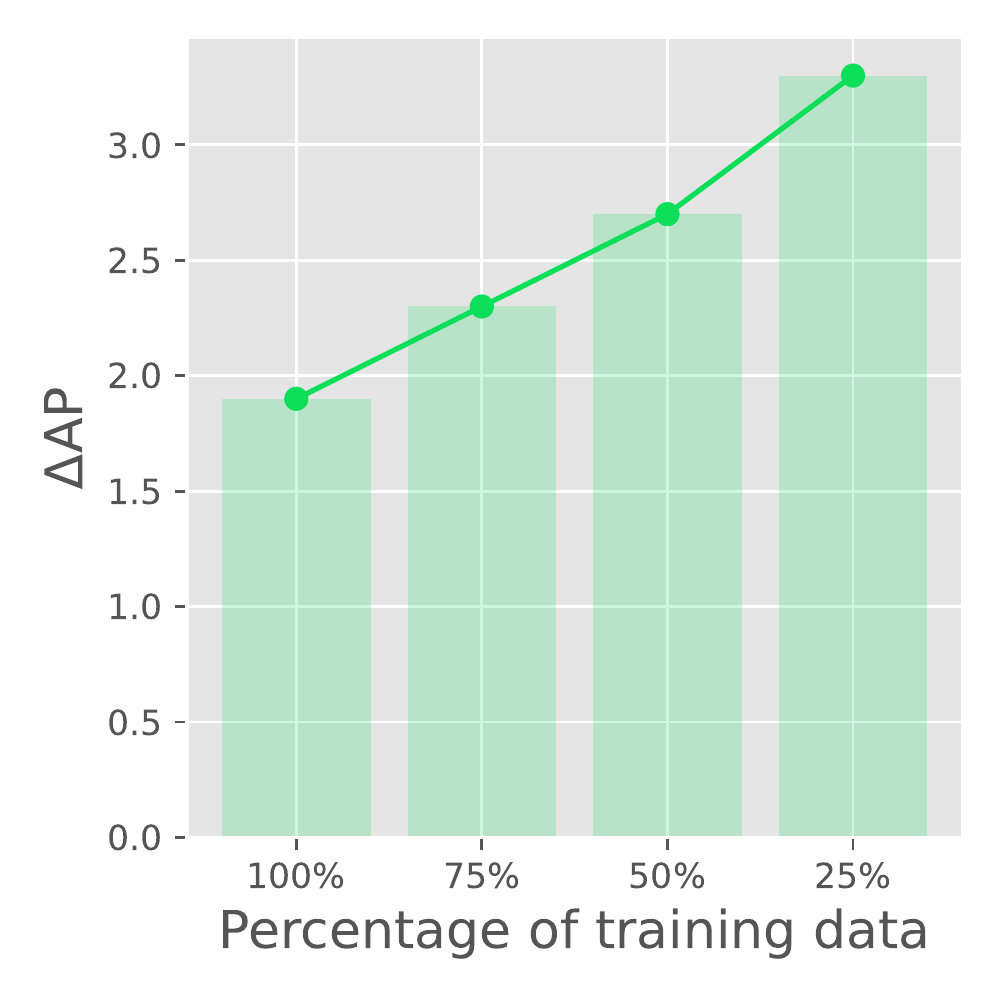}
    \quad
    \includegraphics[width=0.47\linewidth, height=0.5\linewidth]{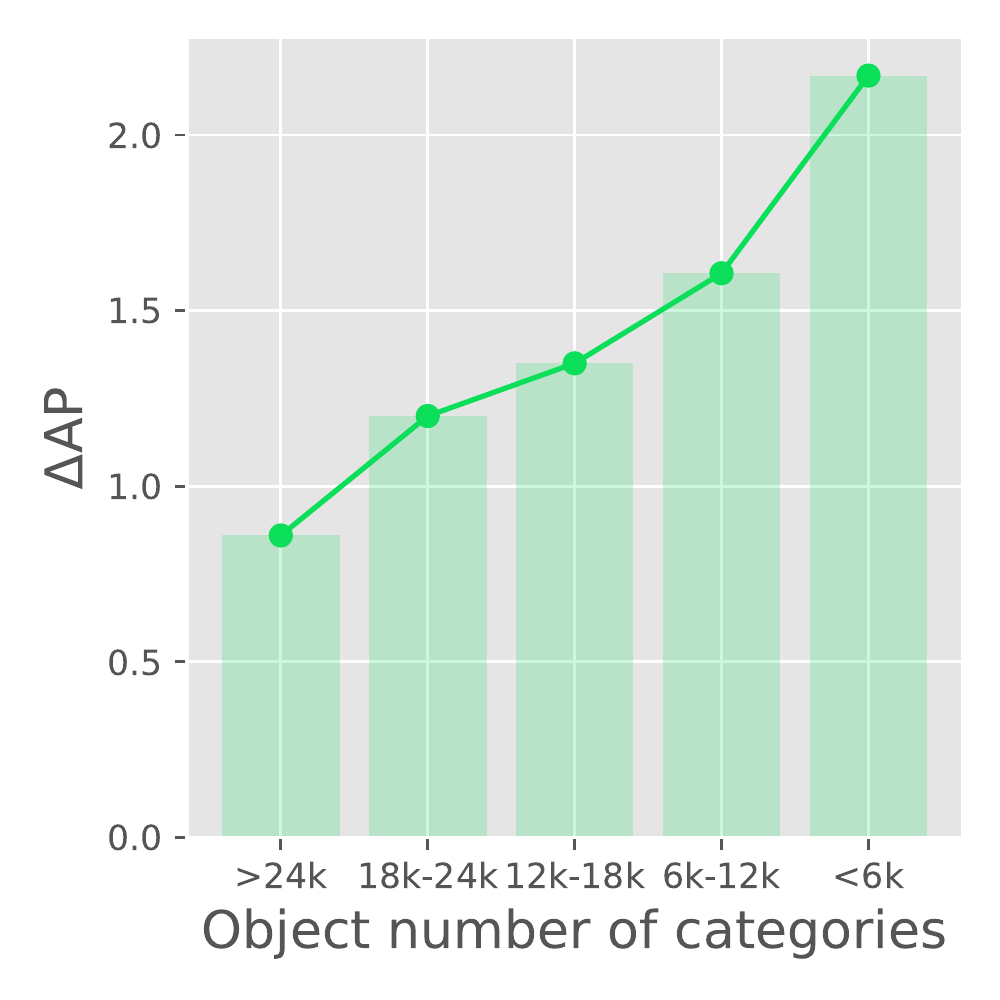}
  \caption{Performance gains on low-shot object detection. Left: Performance gains when reducing training samples. Right: Performance gains with respect to training sample numbers.}
  \label{fig:generalization}
\end{figure}

\textbf{Low-shot Object Detection.} imTED has greater generalization capacity because its feature extraction procedure is consistent with the pre-trained representation models. To validate this capacity, we evaluate the performance gains of imTED over the baseline detector by gradually reducing the training samples, which is termed low-shot object detection, Fig.~\ref{fig:generalization}(left). When the percentage of training data reduces, the performance gains of imTED over the baseline detector monotonously increase. Larger performance gains with less training data demonstrate greater generalization capability.

We also evaluate the detection performance of object categories under different numbers of training instances. As shown in Fig.~\ref{fig:generalization}(right), for the object categories of fewer training instances, imTED outperforms the baseline detector by larger margins. This further validates the effectiveness of imTED for low-shot object detection, which implies higher generalization capability.

\textbf{Few-shot Object Detection.} imTED can be applied for few-shot object detection without any modification. Following Meta YOLO~\cite{kang2019few}, the object categories in MS COCO are divided into two groups: base classes with adequate annotations and novel classes with $K$-shot annotated instances. On MS COCO, 20 classes are selected as novel ones and the remaining 60 classes as base ones. The base classes are used to initialize the detector, $i.e.$, endowing it the ability to localize objects, through base training. The detector is then finetuned upon the novel classes for few-shot object detection. In Table~\ref{tab:few-shot-det}, imTED respectively improves the state-of-the-arts of few-shot detection by 3.5 (19.0 to 22.5) and 7.6 (22.6 to 30.2) under 10-shot and 30-shot settings. 
The large performance gains further validate the generalization capability of the proposed imTED detectors.

\textbf{Occluded Object Detection.}
We configure a sub-set of (534) images with occluded objects from the validation set of MS COCO. If two ground-truth objects has an IoU larger than 0.5, the corresponding image will be selected.
In Table~\ref{tab:occlusion}, by introducing the decoder and removing FPN, imTED improves the AP performance of occluded object detection by 1.2 (36.6 to 37.8). 
When using FPN as the modulator, the AP improvement increases to 2.4 (36.6 to 39.0). 
The total performance gain on the occluded subset is larger than that of the full set of MS COCO (2.4 \vs 1.6), demonstrating the superiority of integral migration on the occluded object detection task.
As is known, MAE learns features via a form of denoising autoencoder, where each image is occluded with random patch masks and fed to the encoder while the decoder predicts the original pixel values of the masked (occluded) patches. 
This occlusion-and-prediction procedure performed on a large mount of images enables MAE models intrinsically learning occlusion invariant features. By integral migration, the imTED detector retains the capacity of MAE pre-trained models, which facilities detecting occluded objects.

\begin{table}[t]
  \caption{
     Performance comparison of few-shot object detection on the MS COCO dataset. }
  \label{tab:few-shot-det}
  \centering
  \small
  \begin{tabular}{lllc}
    \toprule
    Shots  &Method                         &Detector            &AP \\
    \midrule
    \multirow{7}{*}{10} 
    &Meta YOLO~\cite{kang2019few}   &YOLOv2  &5.6 \\
    &CME~\cite{li2021beyond}        &FasterR-CNN + R101    &15.1 \\
    &FCT~\cite{han2022few}          &PVTv2-B2-Li        &17.1  \\
    &Meta-DETR~\cite{zhang2021meta} &DETR + R101          &19.0\\
    &DeFRCN~\cite{qiao2021defrcn}   &FasterR-CNN + R101    &18.5\\
    &Baseline                    &Faster R-CNN + ViT-B        &14.8\\
    %&imTED(ours)                    &imTED + ViT-S        &15.0\\
    &imTED(ours)                    &imTED + ViT-B        &\bf22.5 \\
    %&imTED-FPN(ours)                    &imTED-FPN + ViT-S        &15.6 \\
    % &imTED(ours)                    &imTED + ViT-B        &\\
    \midrule
    \multirow{7}{*}{30} 
    &Meta YOLO~\cite{kang2019few}   &YOLOv2  &9.1 \\
    &CME~\cite{li2021beyond}        &FasterR-CNN + R101    &16.9\\
    &FCT~\cite{han2022few}          &PVTv2-B2-Li        &21.4\\
    &Meta-DETR~\cite{zhang2021meta} &DETR + R101          &22.2 \\
    &DeFRCN~\cite{qiao2021defrcn}   &FasterR-CNN + R101    &22.6\\
    &Baseline                    &Faster R-CNN + ViT-B        &22.2\\
    %&imTED(ours)                    &imTED + ViT-S        &21.0\\
    &imTED(ours)                    &imTED + ViT-B        &\bf30.2\\
    %&imTED-FPN(ours)                    &imTED-FPN + ViT-S        &\\
    % &imTED(ours)                    &imTED + ViT-B        &\\
    \bottomrule
  \end{tabular}
\end{table}

\begin{table}[t]
  \caption{
    Ablation studies using ViT-S as the backbone (encoder) on occluded objects in 1x schedule. \mark~indicates that the module is initialized with MAE pre-trained weights. }
  \label{tab:occlusion}
  \centering
  \resizebox{\linewidth}{!}{ %resize
  \begin{tabular}{lcccccccccc}
    \toprule
    Detector Head   &FPN     & MFM       & AP       & AP\50     & AP\75            \\
    \midrule
    % FC Layers     & \cmark   &  \xmark       & 34.5     & 60.1      & 34.1      \\
    Conv Layers     & \cmark   &  \xmark    & 36.6    & 55.5      & 38.8      \\
    Decoder        & \cmark   &  \xmark       & 36.9     & 56.1      & 39.2     \\
    Decoder\mark      & \cmark   &  \xmark       & 37.5     & 57.1      & 39.5     \\
    Decoder\mark    & \xmark & \xmark      & 37.8     & 57.3      & 42.2      \\
    Decoder\mark     & \xmark & \cmark    & 39.0     & 58.9      & 41.3      \\
    \bottomrule
  \end{tabular}
  }
\end{table}

%-------------------------------------------------------------------------
\section{Conclusion and Future Remarks}
 
We significantly improved the conventional detection pipeline by integrally migrating pre-trained transformer encoder-decoders (imTED). The key idea is to construct a feature extraction path which is not only ``fully pre-trained" but also consistent with MAE models. By migrating an MAE decoder to the detector head and removing FPN, imTED updated Faster R-CNN to a simpler yet more effective detector, where FPN is employed as a feature modulator to further enhance scale adaptability. Experiments on general, low/few-shot and occluded object detection demonstrated the performance gains brought by imTED, with striking contrast with the state-of-the-arts. imTED provides an insight to fully exploit the potential of pre-trained masked autoencoders.

Despite the fact that imTED is implemented with less parameters, the computational cost of the decoder is moderately larger than the detector head with Conv layers. 
In the future, one solution is to use cascaded rejection strategies to reduce object proposals. The other solution is to configure a light-weight decoder using knowledge distillation.

%%%%%%%%% REFERENCES
{\small
\bibliographystyle{ieee_fullname}
\bibliography{egbib}
}

\end{document}